\documentclass[a4paper,twoside]{article}

\usepackage{epsfig}
\usepackage{subcaption}
\usepackage{calc}
\usepackage{amssymb}
\usepackage{amstext}
\usepackage{amsmath}
\usepackage{amsthm}
\usepackage{multicol}
\usepackage{pslatex}
\usepackage{apalike}
\usepackage{algorithm2e}
\usepackage[bottom]{footmisc}

\usepackage{url}

\usepackage{multirow}
\usepackage{adjustbox}
\usepackage{booktabs}
\usepackage{arydshln}
\usepackage[framemethod=TikZ]{mdframed}
\usepackage{pgfplots}
\usetikzlibrary{patterns}
\usepgfplotslibrary{fillbetween}

\usepackage{SCITEPRESS}     

\pgfplotsset{compat=1.18}

\definecolor{lightblue}{HTML}{5DA5DA}
\definecolor{lightorange}{HTML}{FAA43A}
\definecolor{lightgreen}{HTML}{60BD68}
\definecolor{lightpurple}{HTML}{B276B2}
\definecolor{lightgray}{rgb}{0.97,0.97,0.97} 
\definecolor{darkgray}{rgb}{0.8,0.8,0.8} 
\definecolor{lightgreenprompt}{HTML}{90EE90}

\makeatletter
\def\adl@drawiv#1#2#3{%
        \hskip.5\tabcolsep
        \xleaders#3{#2.5\@tempdimb #1{1}#2.5\@tempdimb}%
                #2\z@ plus1fil minus1fil\relax
        \hskip.5\tabcolsep}
\newcommand{\cdashlinelr}[1]{%
  \noalign{\vskip\aboverulesep
           \global\let\@dashdrawstore\adl@draw
           \global\let\adl@draw\adl@drawiv}
  \cdashline{#1}
  \noalign{\global\let\adl@draw\@dashdrawstore
           \vskip\belowrulesep}}
\makeatother

\newcommand{\quotes}[1]{``#1''}
\renewcommand{\vec}[1]{\ensuremath{\mathbf{#1}}}

\usepackage{hyperref}
\definecolor{darkblue}{rgb}{0, 0, 0.5}
\hypersetup{colorlinks=true, citecolor=black, linkcolor=black, urlcolor=darkblue}

\usepackage{fancyhdr}

\fancypagestyle{firstpage}{
  \fancyhf{} 
  \fancyfoot[C]{\thepage} 
  \fancyfoot[L]{\scriptsize Accepted at ICAART 2025. Official version: \href{https://doi.org/10.5220/0013349400003890}{doi.org/10.5220/0013349400003890}} 
}

\begin{document}
\title{Advancing Cross-lingual Aspect-Based Sentiment Analysis with LLMs and Constrained Decoding for Sequence-to-Sequence Models}
\thispagestyle{firstpage} 

\author{\authorname{Jakub \v{S}m\'{i}d\sup{1,2}\orcidAuthor{0000-0002-4492-5481}, Pavel P\v{r}ib\'{a}\v{n}\sup{1}\orcidAuthor{0000-0002-8744-8726} and Pavel Kr\'{a}l\sup{2}\orcidAuthor{0000-0002-3096-675X}}
\affiliation{\sup{1}Department of Computer Science and Engineering, University of West Bohemia in Pilsen, Univerzitn\'{i}, Pilsen, Czech Republic}
\affiliation{\sup{2}NTIS - New Technologies for the Information Society, University of West Bohemia in Pilsen, Univerzitn\'{i}, Pilsen, Czech Republic}
\email{\{jaksmid, pribanp, pkral\}@kiv.zcu.cz}
}


\abstract{
Aspect-based sentiment analysis (ABSA) has made significant strides, yet challenges remain for low-resource languages due to the predominant focus on English. Current cross-lingual ABSA studies often centre on simpler tasks and rely heavily on external translation tools. In this paper, we present a novel sequence-to-sequence method for compound ABSA tasks that eliminates the need for such tools. Our approach, which uses constrained decoding, improves cross-lingual ABSA performance by up to 10\%. This method broadens the scope of cross-lingual ABSA, enabling it to handle more complex tasks and providing a practical, efficient alternative to translation-dependent techniques. Furthermore, we compare our approach with large language models (LLMs) and show that while fine-tuned multilingual LLMs can achieve comparable results, English-centric LLMs struggle with these tasks.
}

\onecolumn \maketitle \normalsize \setcounter{footnote}{0} \vfill


\section{\texorpdfstring{\uppercase{Introduction}}{Introduction}}
Sentiment analysis aims to understand and quantify opinions expressed in text, playing a critical role in applications like customer feedback analysis, social media monitoring, and market research. Within this field, aspect-based sentiment analysis (ABSA) focuses on extracting fine-grained sentiment elements from text~\cite{absa}. These elements include aspect term ($a$), aspect category ($c$), and sentiment polarity ($p$). For example, in the review \textit{\quotes{Tasty soup}}, these elements are \textit{\quotes{soup}}, \textit{\quotes{food quality}}, and \quotes{\textit{positive}}, respectively. Implicitly referenced aspect terms, as in \textit{\quotes{Delicious}}, are frequently labelled as \textit{\quotes{NULL}}.

Initially, ABSA research focused on individual sentiment elements, e.g. aspect term extraction and aspect category detection~\cite{pontiki-etal-2014-semeval}. Recent studies have shifted towards compound tasks involving multiple elements, such as end-to-end ABSA (E2E-ABSA), aspect category term extraction (ACTE), and target aspect category detection (TASD)~\cite{tasd}. Table~\ref{tab:absa-tasks} shows the output formats of these ABSA tasks.

\begin{table}[ht!]
    \caption{Output format for selected ABSA tasks for an input review: \textit{\quotes{Tasty soup but pricey tea}}.}
    \centering
    \begin{adjustbox}{width=0.99\linewidth}
        \begin{tabular}{@{}lll@{}}
            \toprule
            \textbf{Task} &  \textbf{Output}     & \textbf{Example output}  \\                        \midrule
            E2E-ABSA      &  \{($a$, $p$)\}      & \{(\quotes{soup}, POS), (\quotes{tea}, NEG)\}               \\
            ACTE          &  \{($a$, $c$)\}      & \{(\quotes{soup}, food), (\quotes{tea}, drinks)\}           \\
            TASD          & \{($a$, $c$, $p$)\} & \{(\quotes{soup}, food, POS), (\quotes{tea}, drinks, NEG)\} \\ \bottomrule
        \end{tabular}
    \end{adjustbox}
	\label{tab:absa-tasks}
\end{table}

While ABSA research traditionally focuses on English, real-world applications demand multilingual capabilities. However, annotating multilingual data is costly and time-intensive. Although multilingual pre-trained models have become standard for cross-lingual transfer in natural language processing (NLP) tasks~\cite{pmlr-v119-hu20b}, applying them to cross-lingual ABSA presents challenges due to language-specific knowledge requirements. These models are usually fine-tuned on source language data and directly applied to target language data. However, they might lack language-specific knowledge for ABSA tasks involving user-generated texts with abbreviations, slang, and language-dependent aspects. A possible solution is using translated target language data with projected labels, but its effectiveness depends on the quality of the translation and alignment.

Modern monolingual ABSA approaches use pre-trained sequence-to-sequence models, framing compound tasks as text generation problems. In contrast, cross-lingual ABSA research remains limited, focusing mainly on simple tasks and E2E-ABSA, with no studies employing the sequence-to-sequence methods that excel in monolingual ABSA.

Recent advancements in large language models (LLMs), such as GPT-4o~\cite{openai2024gpt4o} and LLaMA~3\cite{llama3modelcard}, have achieved remarkable results across NLP tasks. However, fine-tuned models outperform LLMs on compound ABSA tasks~\cite{zhang-etal-2024-sentiment}. Fine-tuned LLaMA-based models lead in English ABSA~\cite{smid-etal-2024-llama}, but their cross-lingual performance remains unexplored.

\begin{figure*}[ht!]
    \centering
    \includegraphics[width=0.95\linewidth]{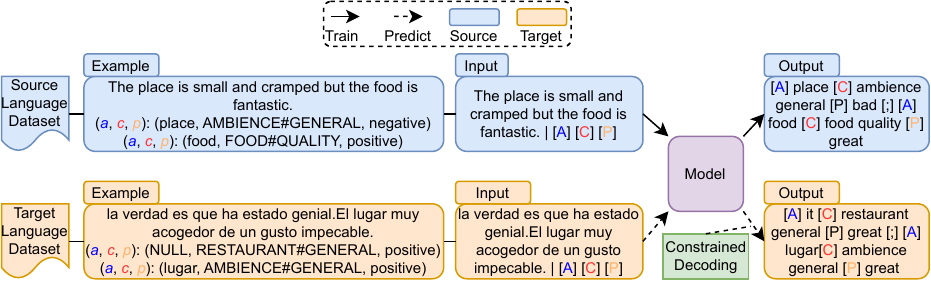}
    \caption{Overview of the proposed framework, which includes converting input labels to natural language phrases, fine-tuning on source language data, and making predictions on target language data using constrained decoding for enhancement.}
    \label{fig:overview}
\end{figure*}

The main motivation of this paper is the limited research on compound cross-lingual ABSA tasks, the absence of sequence-to-sequence approaches widely used in monolingual ABSA, and the reliance on external translation tools in related work, which adds complexity and potential error to the process. To address these shortcomings in existing works, we introduce a novel sequence-to-sequence method that achieves favourable results for compound ABSA tasks in cross-lingual settings without relying on external translation tools. Additionally, we explore the capabilities of several LLMs for cross-lingual ABSA, as their performance on this specific task has not been thoroughly investigated.

Our main contributions are as follows: 1) We introduce the first sequence-to-sequence approach for compound cross-lingual ABSA tasks, which does not rely on external translation tools. 2) We significantly improve zero-shot cross-lingual ABSA performance using constrained decoding. 3) We compare our method to LLMs, specifically GPT-4o mini and fine-tuned LLaMA~3 and LLaMA~3.1, showing that only fine-tuned multilingual LLaMA~3.1 achieves comparable results to our approach. 4) We conduct experiments on benchmark datasets in five languages and three compound ABSA tasks, achieving new state-of-the-art results in both cross-lingual and monolingual settings. To the best of our knowledge, we are the first to examine two compound cross-lingual ABSA tasks and the cross-lingual capabilities of LLMs for ABSA.

\section{\texorpdfstring{\uppercase{Related Work}}{Related Work}}
Cross-lingual ABSA research focuses on three main tasks: aspect term extraction~\cite{klinger-cimiano-2015-instance,wang2018transition,jebbara-cimiano-2019-zero}, aspect sentiment classification~\cite{lambert-2015-aspect,barnes-etal-2016-exploring,akhtar-etal-2018-solving}, and E2E-ABSA~\cite{li2020unsupervised,zhang-etal-2021-cross,lin2023clxabsa,LIN2024125059}. Of these tasks, only E2E-ABSA is compound task, i.e. it focuses on extracting more than one sentiment element simultaneously. Early methods relied on translation and word alignment tools like FastAlign~\cite{dyer-etal-2013-simple}, with quality improvements through instance selection~\cite{klinger-cimiano-2015-instance} or constrained translation~\cite{lambert-2015-aspect}. Others used cross-lingual embeddings trained on bilingual corpora for language-independent ABSA~\cite{lambert-2015-aspect,barnes-etal-2016-exploring,akhtar-etal-2018-solving,wang2018transition,jebbara-cimiano-2019-zero}. Recent work focus on multilingual Transformer-based~\cite{vaswani2017attention} encoder-only models, such as mBERT~\cite{devlin-etal-2019-bert} and XLM-R~\cite{conneau-etal-2020-unsupervised} combined with machine translation, with additional enhancements from parameter warm-up~\cite{li2020unsupervised}, distillation on unlabelled target language data~\cite{zhang-etal-2021-cross}, contrastive learning~\cite{lin2023clxabsa}, and dynamic loss weighting~\cite{LIN2024125059}.

The latest monolingual ABSA research focuses primarily on text generation, exploring converting labels to natural language~\cite{zhang-etal-2021-towards-generative,zhang-etal-2021-aspect-sentiment}, multi-tasking~\cite{gao-etal-2022-lego}, generating tuples as paths of a tree~\cite{mao-etal-2022-seq2path}, element ordering~\cite{hu-etal-2022-improving-aspect,gou-etal-2023-mvp}, and tagging-assisted generation~\cite{xianlong-etal-2023-tagging}.

Research shows that fine-tuned models outperform non-fine-tuned LLMs in compound ABSA tasks~\cite{gou-etal-2023-mvp,zhang-etal-2024-sentiment}, whereas fine-tuned LLaMA-based models achieve state-of-the-art results in English ABSA~\cite{smid-etal-2024-llama}.

\section{\texorpdfstring{\uppercase{Methodology}}{Methodology}}
This section presents our approach to addressing the triplet task (TASD), which can be easily modified for tuple tasks with minor adjustments. Figure~\ref{fig:overview} depicts the proposed approach.

\subsection{Problem Definition}
Given an input sentence, the aim is to predict all sentiment tuples $T={(a, c, p)}$, each composed of an aspect term ($a$), aspect category ($c$), and sentiment polarity ($p$). Following prior works~\cite{zhang-etal-2021-aspect-sentiment,gou-etal-2023-mvp}, we convert elements ($a, c, p$) into natural language ($e_a, e_c, e_p$). For instance, we translate the \quotes{\textit{neutral}} sentiment polarity into \textit{\quotes{ok}} and the \textit{\quotes{NULL}} aspect term into \textit{\quotes{it}}, as shown in Figure~\ref{fig:overview}.

\subsection{Input and Output Building}

To build our model's inputs and outputs, we employ element markers to represent each sentiment element: {\sffamily[A]} for $e_a$, {\sffamily[C]} for $e_c$, and {\sffamily[P]} for $e_p$. These markers prefix each element, forming the target sequence together. We also append these markers to the input sequence to guide the model for correct output. We follow the priority order $e_a > e_c > e_p$ recommended in prior research~\cite{gou-etal-2023-mvp}. For example, we create the following input-output pair: \\
\hspace*{7pt}\textbf{\textit{Input ($x$)}:} They offer a tasty soup \text{\textbar} {\sffamily[A]} {\sffamily[C]} {\sffamily[P]} \\
\hspace*{7pt}\textbf{\textit{Output ($y$)}:} {\sffamily[A]} soup {\sffamily[C]} food quality {\sffamily[P]} great

For sentences with multiple sentiment tuples, we use the symbol {\sffamily[;]} to concatenate their target schemes into the final target sequence. Different examples are depicted in Figure~\ref{fig:overview}.

\subsection{Constrained Decoding}

To prevent the fine-tuned model from generating aspect terms in the source language instead of the target language, we have developed scheme-guided constrained decoding (CD)~\cite{constrained}, which ensures that generated elements match their respective vocabulary sets by incorporating target schema information. This approach is beneficial in few-shot monolingual settings~\cite{gou-etal-2023-mvp}. 
\begin{table}[ht!]
    \caption{Candidate lists of tokens. \texttt{<eos>} indicates the end of a sequence, and \quotes{$\ldots$} denotes arbitrary text.}
    \centering
        \begin{tabular}{@{}ll@{}}
            \toprule
            \textbf{Generated output}                & \textbf{Candidate tokens} \\ \midrule
                                                    & [                         \\
            $\ldots$ {\sffamily [A} / {\sffamily [C} / {\sffamily [P} / {\sffamily [;} & {\sffamily ]}                         \\
            $\ldots$ {\sffamily [A]}                                   & Input sentence, \quotes{it}            \\
            $\ldots$ {\sffamily [C]}                                   & All categories            \\
            $\ldots$ {\sffamily [P]}                                   & great, ok, bad            \\
            $\ldots$ {\sffamily [A]} $\ldots$                                 & Input sentence, \quotes{it}, {\sffamily [}         \\
            $\ldots$ {\sffamily [C]} $\ldots$                                 & All categories, {\sffamily [}         \\
            $\ldots$ {\sffamily [P]} $\ldots$                                 & great, ok, bad, \texttt{<eos>} {\sffamily [}         \\
            $\ldots$ {\sffamily [A]} $\ldots$ {\sffamily [}                               & {\sffamily C}                         \\
            $\ldots$ {\sffamily [C]} $\ldots$ {\sffamily [}                               & {\sffamily P}                         \\
            $\ldots$ {\sffamily [P]} $\ldots$ {\sffamily [}                               & ;                         \\
            $\ldots$ {\sffamily [;]}                                   & {\sffamily [}                         \\
            $\ldots$ {\sffamily [;]} {\sffamily [}                                 & {\sffamily A}                         \\ \bottomrule
        \end{tabular}
    \label{tab:CD}
\end{table}

Constrained decoding dynamically adjusts candidate token lists based on the current state, enhancing control and accuracy in the generation process. For example, if the current token is '{\sffamily[}', the next token should be chosen from special terms: {\sffamily A}, {\sffamily C}, {\sffamily P}, and {\sffamily ;}. Additionally, it tracks previously generated output and current terms, guiding the decoding of subsequent tokens based on Table~\ref{tab:CD}. Appendix~\ref{appendix:cd} shows the algorithm in more detail.

\subsection{Training}

We fine-tune a pre-trained sequence-to-sequence model with provided input-output pairs. Sequence-to-sequence models consist of two components: the encoder, which transforms input sequence $x$ into a contextualized sequence $\vec{e}$, and the decoder, which models the conditional probability distribution $P_{{\Theta}}(y|\vec{e})$ of the target sequence $y$ based on the encoded input $\vec{e}$, where ${\Theta}$ represents the model's parameters. At each decoding step $i$, the decoder generates the output $y_i$ using previous outputs $y_0, \ldots, y_{i-1}$ and the encoded input $\vec{e}$. During fine-tuning, we update all model parameters and minimize the log-likelihood as
\begin{equation}
    \mathcal{L} = -\sum_{i=1}^n\log p_{{\Theta}}(y_i|\vec{e},y_{<i}),
\end{equation}
where $n$ is the length of the target sequence $y$.

\section{\texorpdfstring{\uppercase{Experiments}}{Experiments}}
\label{sec:experiments}

\begin{table*}[ht!]
    \caption{Dataset statistics for each language. POS, NEG and NEU denote the number of positive, negative and neutral examples, respectively.}
    \centering
    \begin{adjustbox}{width=0.8\linewidth}
        \begin{tabular}{@{}llrrrrrr@{}}
            \toprule
                                   &              & \textbf{En}  & \textbf{Es}   & \textbf{Fr}     & \textbf{Nl} & \textbf{Ru}   & \textbf{Tr} \\ \midrule
            \multirow{5}{*}{Train} & Sentences    & 1,800        & 1,863         & 1,559           & 1,549       & 3,289         & 1,108       \\
                                   & Triplets     & 2,266        & 2,455         & 2,276           & 1,676       & 3,697         & 1,386       \\
                                   & Categories   & 12           & 12            & 12              & 12          & 12            & 12          \\
                                   & POS/NEG/NEU  & 1,503/672/91 & 1,736/607/112 & 1,045/1,092/139 & 969/584/124 & 2,805/641/250 & 746/521/119 \\
                                   & NULL aspects & 569          & 700           & 694             & 513         & 821           & 135         \\ \cdashlinelr{1-8}
            \multirow{5}{*}{Dev}   & Sentences    & 200          & 207           & 174             & 173         & 366           & 124         \\
                                   & Triplets     & 241          & 265           & 254             & 184         & 392           & 149         \\
                                   & Categories   & 11           & 11            & 12              & 11          & 12            & 10          \\
                                   & POS/NEG/NEU  & 154/77/10    & 189/67/8      & 115/120/15      & 94/62/28    & 298/68/26     & 74/65/10    \\
                                   & NULL aspects & 58           & 83            & 66              & 64          & 109           & 15          \\ \cdashlinelr{1-8}
            \multirow{5}{*}{Test}  & Sentences    & 676          & 881           & 694             & 575         & 1,209         & 144         \\
                                   & Triplets     & 859          & 1,072         & 954             & 613         & 1,300         & 159         \\
                                   & Categories   & 12           & 12            & 13              & 13          & 12            & 11          \\
                                   & POS/NEG/NEU  & 611/204/44   & 750/274/48    & 441/434/79      & 369/211/33  & 870/321/103   & 104/49/6    \\
                                   & NULL aspects & 209          & 341           & 236             & 219         & 325           & 0           \\ \bottomrule
        \end{tabular}
    \end{adjustbox}
    \label{tab:data_stats}
\end{table*}

We report results with a 95\% confidence interval from 5 runs with different seeds. The primary evaluation metric is the micro F1-score, the standard metric in ABSA research. We consider a predicted sentiment tuple correct only if all its elements exactly match the gold tuple.

\subsection{{Tasks and Dataset}}
We evaluate our method on the E2E-ABSA, ACTE, and TASD tasks (see Table~\ref{tab:absa-tasks} for task details).

We perform experiments on the standard SemEval-2016 dataset~\cite{pontiki-etal-2016-semeval} with restaurant reviews in English (en), Spanish (es), French (fr), Dutch (nl), Russian (ru), and Turkish (tr), with provided training and test sets. We split the training data into a 9:1 ratio to create a validation set. We consider English as the source language and other languages as the target ones. Table~\ref{tab:data_stats} shows the data statistics for each language.

\begin{figure*}[ht!]
    \centering
    \includegraphics[width=0.95\linewidth]{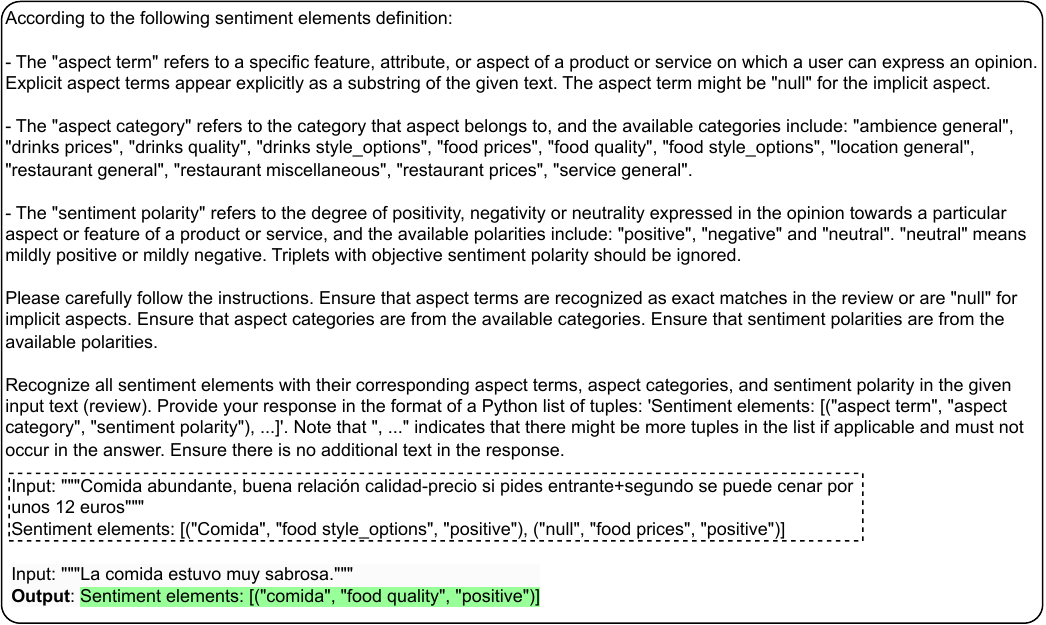}
    \caption{Prompt for the TASD task with example input, expected output in a green box, and one demonstration in Spanish enclosed in a dashed box. The demonstrations are used solely in few-shot scenarios.}
    \label{fig:prompt}
\end{figure*}

\subsection{Prompts for LLMs}
Figure~\ref{fig:prompt} shows the prompt for LLMs for the TASD task, including one example for few-shot settings for Spanish. This prompt can be adapted for various tasks by excluding the unnecessary sentiment element for the specific task, such as the sentiment polarity for the ACTE task. For few-shot prompts, examples are drawn from the first 10 examples of the training dataset in the respective language.

\subsection{Compared Methods}
We compare our method with constrained decoding against models without it. For E2E-ABSA—the only task with available related work—we evaluate against approaches using decoder-only models. One such method~\cite{li2020unsupervised} operates in a true zero-shot setting, training only on source language data without machine translation. Other methods integrate machine translation with additional enhancements, including alignment-free projection and aspect code-switching for interchanging aspect terms between languages with distillation on unlabelled target language data~\cite{zhang-etal-2021-cross}, contrastive learning~\cite{lin2023clxabsa}, and dynamically weighted loss combined with anti-decoupling to improve semantic information utilization and address class imbalances~\cite{LIN2024125059}.

It is important to note that these methods use slightly different task definitions and datasets. Specifically, previous work employs decoder-only models, excludes implicit aspect terms (\textit{\quotes{NULL}}), and restricts each aspect term to a single sentiment polarity. For example, prior research~\cite{zhang-etal-2021-cross} reports 612 tuples in the English test set after filtering \textit{\quotes{NULL}} aspect terms and merging sentiment polarities for each aspect term, whereas we report 859 tuples. In contrast, our approach predicts \textit{\quotes{NULL}} aspect terms and allows multiple sentiment polarities per aspect term, making the task inherently more challenging.

\subsection{Experimental Details}
We use the large mT5~\cite{xue-etal-2021-mt5}, selected based on related work for English~\cite{zhang-etal-2021-towards-generative,zhang-etal-2021-aspect-sentiment,gao-etal-2022-lego,gou-etal-2023-mvp} employing monolingual T5~\cite{raffel2020exploring}, and the large mBART~\cite{tang2020multilingual} to evaluate our method across different architectures. Both models are from the HuggingFace Transformers library\footnote{\url{https://github.com/huggingface/transformers}}~\cite{wolf-etal-2020-transformers}. We fine-tune the models for all experiments over 20 epochs with a batch size of 16 and employ greedy search for decoding. For mT5, we use a learning rate of 1e-4 and the Adafactor optimizer~\cite{adafactor}. For mBART, we use a learning rate of 1e-5 and the AdamW optimizer~\cite{loshchilov2017decoupled}. These settings were chosen based on consistent performance on validation data across all languages and tasks.

We evaluate GPT-4o mini~\cite{openai2024gpt4o} using zero- and few-shot prompts.
Additionally, we fine-tune the 8B versions of LLaMA~3~\cite{llama3modelcard} and LLaMA~3.1~\cite{dubey2024llama3herdmodels}, employing QLoRA~\cite{dettmers2023qlora} with 4-bit NormalFloat quantization, a batch size of 16, a constant learning rate of 2e-4, AdamW optimizer, LoRA adapters~\cite{hu2021lora} on all linear Transformer block layers, and LoRA $r=64$ and $\alpha=16$. Utilizing the zero-shot prompt shown in Figure~\ref{fig:prompt}, i.e. without the demonstrations, we fine-tune the model for up to 5 epochs, selecting the best-performing model based on validation loss. All experiments are conducted using an NVIDIA A40 with 48 GB GPU.

\section{\texorpdfstring{\uppercase{Results}}{Results}}

\begin{table*}[ht!]
\caption{F1 scores for zero-shot cross-lingual ABSA with English as the source language and other languages as target languages compared to monolingual results and GPT-4o mini. The compared works have different models and E2E-ABSA definitions. \textbf{Bold} results indicate significant improvements using constrained decoding (CD). \underline{Underlined} results are the best absolute results for each language and task in both monolingual and cross-lingual settings.} 
\centering
\begin{adjustbox}{width=\linewidth}
\begin{tabular}{@{}lllccccccccccccccc@{}}
\toprule
                               &                                &                  & \multicolumn{5}{c}{\textbf{E2E-ABSA}}                                                                                                                   & \multicolumn{5}{c}{\textbf{ACTE}}                                                                                                                  & \multicolumn{5}{c}{\textbf{TASD}}                                                                                                                   \\ \cmidrule(lr){4-8} \cmidrule(lr){9-13} \cmidrule(lr){14-18}
                               &                                &                  & Es                        & Fr                        & Nl                        & Ru                        & Tr                        & Es                        & Fr                        & Nl                        & Ru                        & Tr                        & Es                        & Fr                        & Nl                        & Ru                        & Tr                         \\ \midrule
\multirow{8}{*}{\rotatebox[origin=c]{90}{Monolingual}}   & \multirow{2}{*}{mT5}           & w/o CD       & 74.4$^{\pm0.6}$           & 69.9$^{\pm0.5}$           & 71.6$^{\pm1.0}$           & 72.4$^{\pm0.2}$           & 60.1$^{\pm1.7}$           & 70.4$^{\pm0.7}$           & 63.7$^{\pm0.8}$           & 68.8$^{\pm0.5}$           & 73.2$^{\pm0.5}$           & 59.1$^{\pm0.5}$           & 65.8$^{\pm0.4}$           & 59.0$^{\pm0.6}$           & 62.9$^{\pm1.4}$           & 67.0$^{\pm0.9}$           & 54.1$^{\pm3.0}$            \\
                               &                                & w/ CD          & 75.3$^{\pm0.6}$           & 69.8$^{\pm1.4}$           & 67.0$^{\pm0.4}$           & 72.2$^{\pm0.4}$           & \underline{60.7}$^{\pm1.1}$           & 69.9$^{\pm0.4}$           & 64.9$^{\pm0.5}$           & 62.9$^{\pm0.5}$           & 72.8$^{\pm1.0}$           & 60.4$^{\pm2.1}$           & \underline{66.2}$^{\pm0.5}$           & 58.9$^{\pm1.1}$           & 57.6$^{\pm0.5}$           & 66.4$^{\pm0.4}$           & 53.9$^{\pm1.5}$            \\ \cdashlinelr{2-18}
                               & \multirow{2}{*}{mBART}         & w/o CD       & 73.0$^{\pm0.5}$           & 66.4$^{\pm1.1}$           & 68.9$^{\pm1.2}$           & 68.7$^{\pm1.6}$           & 56.0$^{\pm2.7}$           & 66.4$^{\pm1.6}$           & 61.1$^{\pm1.6}$           & 64.1$^{\pm1.2}$           & 70.9$^{\pm0.6}$           & 56.8$^{\pm2.2}$           & 62.9$^{\pm1.2}$           & 54.8$^{\pm0.9}$           & 57.6$^{\pm0.9}$           & 62.6$^{\pm0.7}$           & 49.3$^{\pm3.1}$             \\
                               &                                & w/ CD          & 71.9$^{\pm1.3}$           & 64.0$^{\pm1.7}$           & 61.6$^{\pm1.0}$           & 66.2$^{\pm1.1}$           & 54.4$^{\pm2.3}$           & 66.8$^{\pm1.5}$           & 58.2$^{\pm1.2}$           & 58.0$^{\pm1.2}$           & 67.4$^{\pm0.3}$           & 55.3$^{\pm1.5}$           & 61.5$^{\pm1.4}$           & 52.4$^{\pm0.6}$           & 52.1$^{\pm1.0}$           & 60.1$^{\pm1.9}$           & 47.6$^{\pm2.7}$             \\ \cdashlinelr{2-18}
                               & \multirow{2}{*}{GPT-4o mini}       & 0-shot        & 55.7\phantom{$^{\pm0.0}$} & 52.5\phantom{$^{\pm0.0}$} & 45.8\phantom{$^{\pm0.0}$} & 48.3\phantom{$^{\pm0.0}$} & 47.9\phantom{$^{\pm0.0}$} & 34.8\phantom{$^{\pm0.0}$} & 31.7\phantom{$^{\pm0.0}$} & 27.9\phantom{$^{\pm0.0}$} & 31.7\phantom{$^{\pm0.0}$} & 27.4\phantom{$^{\pm0.0}$} & 32.4\phantom{$^{\pm0.0}$} & 31.8\phantom{$^{\pm0.0}$} & 29.1\phantom{$^{\pm0.0}$} & 31.0\phantom{$^{\pm0.0}$} & 32.2\phantom{$^{\pm0.0}$}  \\
                               &                                & 10-shot         & 63.8\phantom{$^{\pm0.0}$} & 55.2\phantom{$^{\pm0.0}$} & 58.9\phantom{$^{\pm0.0}$} & 58.9\phantom{$^{\pm0.0}$} & 51.0\phantom{$^{\pm0.0}$} & 56.2\phantom{$^{\pm0.0}$} & 46.1\phantom{$^{\pm0.0}$} & 49.5\phantom{$^{\pm0.0}$} & 45.8\phantom{$^{\pm0.0}$} & 47.5\phantom{$^{\pm0.0}$} & 50.9\phantom{$^{\pm0.0}$} & 40.9\phantom{$^{\pm0.0}$} & 46.0\phantom{$^{\pm0.0}$} & 42.1\phantom{$^{\pm0.0}$} & 44.3\phantom{$^{\pm0.0}$}  \\ \cdashlinelr{2-18}
                               & \multicolumn{2}{l}{LLaMA 3}                                         & 70.0$^{\pm2.0}$           & 63.1$^{\pm1.8}$           & 66.0$^{\pm1.2}$           & 60.7$^{\pm2.3}$           & 48.6$^{\pm2.0}$           & 59.8$^{\pm2.9}$           & 54.9$^{\pm1.2}$           & 58.1$^{\pm4.3}$           & 62.1$^{\pm1.7}$           & 46.3$^{\pm2.9}$           & 57.2$^{\pm1.7}$           & 48.2$^{\pm2.4}$           & 55.4$^{\pm2.8}$           & 53.7$^{\pm2.9}$         
                               & 39.1$^{\pm3.2}$             \\
                               & \multicolumn{2}{l}{LLaMA 3.1}                                  & \underline{77.4}$^{\pm0.5}$           & \underline{71.2}$^{\pm0.5}$           & \underline{74.0}$^{\pm0.5}$           & \underline{73.7}$^{\pm0.3}$           & 59.2$^{\pm1.6}$           & \underline{70.9}$^{\pm0.6}$           & \underline{66.8}$^{\pm2.3}$           & \underline{70.6}$^{\pm0.8}$           & \underline{75.3}$^{\pm0.9}$           & \underline{60.7}$^{\pm0.5}$           & 65.8$^{\pm1.2}$           & \underline{62.0}$^{\pm0.6}$           & \underline{65.7}$^{\pm1.1}$           & \underline{68.4}$^{\pm1.0}$           & \underline{58.6}$^{\pm1.0}$           \\
                               \midrule
\multirow{10}{*}{\rotatebox[origin=c]{90}{Cross-lingual}} & \multirow{2}{*}{mT5}           & w/o CD       & 59.2$^{\pm0.5}$           & 57.8$^{\pm1.2}$           & 57.1$^{\pm0.9}$           & 56.4$^{\pm2.1}$           & 44.4$^{\pm1.4}$           & 52.5$^{\pm1.0}$           & 55.8$^{\pm0.7}$           & 52.3$^{\pm1.3}$           & 55.0$^{\pm2.7}$           & 41.4$^{\pm1.4}$           & 48.3$^{\pm0.5}$           & 50.4$^{\pm1.4}$           & 47.7$^{\pm1.1}$           & 48.6$^{\pm2.0}$           & 39.1$^{\pm3.6}$             \\
                               &                                &  w/ CD                & \textbf{69.3}$^{\pm1.0}$  & \textbf{61.1}$^{\pm1.2}$  & \textbf{60.8}$^{\pm0.3}$  & \textbf{\underline{63.7}}$^{\pm1.3}$  & \textbf{\underline{48.9}}$^{\pm1.4}$  & \textbf{62.8}$^{\pm1.4}$  & \textbf{57.5}$^{\pm0.3}$  & \textbf{54.1}$^{\pm0.2}$  & \textbf{\underline{60.4}}$^{\pm0.9}$  & \textbf{\underline{49.0}}$^{\pm0.9}$  & \textbf{57.6}$^{\pm0.6}$  & 50.4$^{\pm0.8}$           & \textbf{50.4}$^{\pm1.3}$  & \textbf{\underline{54.9}}$^{\pm2.0}$  & \textbf{\underline{43.8}}$^{\pm0.8}$    \\ \cdashlinelr{2-18}
                               & \multirow{2}{*}{mBART}         & w/o CD       & 61.1$^{\pm2.6}$           & 49.4$^{\pm3.8}$           & 51.6$^{\pm2.7}$           & 57.1$^{\pm1.4}$           & 31.6$^{\pm3.9}$           & 52.5$^{\pm1.4}$           & 49.3$^{\pm1.5}$           & 44.5$^{\pm1.4}$           & 53.8$^{\pm1.5}$           & 31.1$^{\pm2.1}$           & 47.6$^{\pm1.9}$           & 39.6$^{\pm0.8}$           & 39.1$^{\pm0.9}$           & 48.5$^{\pm1.1}$           & 23.5$^{\pm2.6}$             \\
                               &                                & w/ CD          & 61.7$^{\pm2.7}$           & 49.2$^{\pm4.1}$           & 50.1$^{\pm3.5}$           & 57.8$^{\pm1.8}$           & 30.3$^{\pm3.0}$           & \textbf{54.8}$^{\pm0.4}$  & 49.2$^{\pm0.6}$           & \textbf{46.9}$^{\pm0.9}$  & \textbf{55.9}$^{\pm0.2}$  & \textbf{34.7}$^{\pm1.1}$  & \textbf{51.1}$^{\pm1.2}$  & 39.9$^{\pm0.6}$           & 38.9$^{\pm0.9}$           & \textbf{50.5}$^{\pm0.7}$  & \textbf{27.3}$^{\pm1.1}$   \\ \cdashlinelr{2-18}
                               & \multicolumn{2}{l}{LLaMA 3}                                         & 47.2$^{\pm5.0}$           & 41.7$^{\pm3.0}$           & 43.3$^{\pm3.5}$           & 53.1$^{\pm2.2}$           & 26.6$^{\pm9.7}$           & 44.8$^{\pm8.5}$           & 38.1$^{\pm3.5}$           & 40.0$^{\pm6.0}$           & 50.6$^{\pm4.1}$           & 29.9$^{\pm7.1}$           & 39.5$^{\pm7.5}$           & 31.5$^{\pm3.2}$           & 29.8$^{\pm5.7}$           & 46.5$^{\pm2.6}$           & 22.8$^{\pm7.2}$             \\
                               & \multicolumn{2}{l}{LLaMA 3.1}                                  & \underline{73.4}$^{\pm0.7}$           & \underline{68.1}$^{\pm0.5}$           & \underline{64.2}$^{\pm0.8}$           & 58.8$^{\pm1.0}$           & 48.1$^{\pm4.1}$           & \underline{64.9}$^{\pm1.1}$           & \underline{60.9}$^{\pm3.0}$           & \underline{55.7}$^{\pm1.5}$           & 59.2$^{\pm1.6}$           & 47.7$^{\pm2.1}$           & \underline{61.3}$^{\pm0.6}$           & \underline{57.3}$^{\pm1.2}$           & \underline{54.8}$^{\pm0.8}$           & 53.1$^{\pm1.2}$           & 41.4$^{\pm0.7}$           \\
                               \cdashlinelr{2-18}
                               & \multicolumn{2}{l}{\cite{li2020unsupervised}}    & 67.1\phantom{$^{\pm0.0}$} & 56.4\phantom{$^{\pm0.0}$} & 59.0\phantom{$^{\pm0.0}$} & 56.8\phantom{$^{\pm0.0}$} & 46.2\phantom{$^{\pm0.0}$} & -                         & -                         & -                         & -                         & -                         & -                         & -                         & -                         & -                         & -                           \\
                               & \multicolumn{2}{l}{\cite{zhang-etal-2021-cross}} & 69.2\phantom{$^{\pm0.0}$} & 61.0\phantom{$^{\pm0.0}$} & 63.7\phantom{$^{\pm0.0}$} & 62.0\phantom{$^{\pm0.0}$} & -                         & -                         & -                         & -                         & -                         & -                         & -                         & -                         & -                         & -                         & -                           \\
                               & \multicolumn{2}{l}{\cite{lin2023clxabsa}}        & 61.6\phantom{$^{\pm0.0}$} & 49.5\phantom{$^{\pm0.0}$} & 51.0\phantom{$^{\pm0.0}$} & 50.8\phantom{$^{\pm0.0}$} & -                         & -                         & -                         & -                         & -                         & -                         & -                         & -                         & -                         & -                         & -                           \\
                               & \multicolumn{2}{l}{\cite{LIN2024125059}}        & 69.6\phantom{$^{\pm0.0}$} & 60.7\phantom{$^{\pm0.0}$} & 61.3\phantom{$^{\pm0.0}$} & 62.3\phantom{$^{\pm0.0}$} & -                         & -                         & -                         & -                         & -                         & -                         & -                         & -                         & -                         & -                         & -                           \\
                               \bottomrule 
\end{tabular}
\end{adjustbox}
\label{tab:res}
\end{table*}

Table~\ref{tab:res} presents the results.
Some key observations include the following: \\
\hspace*{7pt} 1) Constrained decoding significantly improves cross-lingual ABSA by up to 10\% over baseline models, effectively mitigating the issue where the model predicts aspect terms in English instead of the target language (e.g. \textit{\quotes{place}} instead of \textit{\quotes{lugar}}). The improvement is most noticeable in Spanish and Russian. Constrained decoding is unnecessary in monolingual experiments as the problem does not occur. \\
\hspace*{7pt} 2) MT5 generally outperforms mBART and benefits more from constrained decoding.
\\
\hspace*{7pt} 3) GPT-4o mini performs relatively well for the E2E-ABSA task, which excludes aspect categories. However, its performance is notably lower on the ACTE and TASD tasks, indicating that identifying aspect categories is challenging for the model. Incorporating few-shot prompts boosts performance by 10–20\% in most cases.
\\
\hspace*{7pt} 4) Our approach consistently outperforms GPT-4o mini across all tasks and languages. Notably, we achieve approximately a 20\% improvement on the TASD task across all languages in zero-shot cross-lingual settings. Even when GPT-4o mini is enhanced with 10-shot prompts, our results remain significantly superior. The biggest difference is generally for Russian and French.
\\
\hspace*{7pt} 5) Our method consistently exceeds the performance of fine-tuned LLaMA~3 across all tasks and language combinations, likely due to LLaMA~3 being primarily pre-trained for English.
\\
\hspace*{7pt} 6) LLaMA~3.1 extends language support to include Spanish and French but lacks compatibility with Russian, Dutch and Turkish. It achieves the best results in most tasks and languages in monolingual settings, although mT5 performs similarly when accounting for confidence intervals. In cross-lingual scenarios, mT5 with constrained decoding usually outperforms LLaMA~3.1 for Russian and Turkish and demonstrates similar or slightly inferior performance for other languages. While LLaMA~3.1 does not officially support Dutch, it likely benefits from linguistic similarities with its supported languages.
\\
\hspace*{7pt} 7) Results for Turkish are consistently worse than for other languages, which might be because it is the the only language not from the Indo-European family.

As mentioned in Section~\ref{sec:experiments}, comparing our E2E-ABSA results with prior research is challenging due to methodological differences. Previous works exclude implicit aspect terms, limit one sentiment polarity per aspect term, and use decoder-only models. Moreover, prior approaches rely on external translation tools and can be affected by the translation quality, whereas our method avoids external tools entirely. In contrast, our approach predicts implicit aspect terms, allows multiple sentiment polarities per aspect term, and avoids external tools entirely. Despite these challenges, our method with constrained decoding achieves comparable results and proves. We find constrained decoding to be more practical than relying on external translation tools.

No prior research exists for ABSA tasks beyond E2E-ABSA in cross-lingual settings, which serves as one of the primary motivations for this paper.

\subsection{{Inference and Training Speed}}
Table~\ref{tab:speed} summarizes the average training time per epoch and inference time per example (both absolute in seconds and relative) for various models on the TASD task, with English as the source language and Spanish as the target language.

\begin{table}[ht!]
     \caption{Average absolute and relative training time per epoch and inference time per example for different models on the TASD task, with English as the source language and Spanish as the target language.}
    \centering
   \begin{adjustbox}{width=0.95\linewidth}
        \begin{tabular}{@{}lrrrr@{}}
            \toprule
            \multirow{2}{*}{\textbf{Model}} & \multicolumn{2}{l}{\textbf{Training Time Per Epoch}} & \multicolumn{2}{l}{\textbf{Inference Time Per Example}} \\ \cmidrule(lr){2-3}  \cmidrule(lr){4-5}
                                            & Absolute [s]                  & Relative                 & Absolute [s]                   & Relative                   \\ \midrule
            mT5                             & 210                           & 1.00                     & 0.24                           & 1.00                       \\
            LLaMA~3.1               & 924                          & 4.40                     & 1.04                           & 4.33                                             \\ \bottomrule
        \end{tabular}
    \end{adjustbox}
    \label{tab:speed}
\end{table}

The mT5 model is the reference, with a relative training time of 1.00. The LLaMA~3.1 model is significantly slower, requiring 4.40 times the training time of mT5 and a much higher inference time, 15.32 times that of mT5. This comparison indicates that while larger models like LLaMA~3.1 may offer performance gains, they come with substantial computational costs during both training and inference.

\subsection{Recommendations}
In summary, constrained decoding improves cross-lingual results significantly. Overall, we recommend using the mT5 model with constrained decoding in most scenarios, while LLaMA~3.1 is preferable for languages it specifically supports. The mT5 model consistently outperforms the English-specific LLaMA~3 across all tasks and languages. Moreover, mT5 delivers excellent results compared to the multilingual LLaMA~3.1 despite having approximately seven times fewer parameters while also requiring less training time per epoch and offering faster inference. Additionally, fine-tuning LLaMA models on consumer GPUs demands specialized techniques, which can be a limiting factor. However, LLaMA~3.1 may be the preferred choice if hardware resources are sufficient and training time and inference are not major concerns.

\subsection{Error Analysis}

\begin{figure}[ht!]
    \centering
    \begin{adjustbox}{width=0.95\linewidth}
        \begin{tikzpicture}
            \begin{axis}[
                ybar,
                bar width=9pt,
                xtick={0,1,2},
                ytick={0,10,20,30,40,50,60},
                xticklabels={\scriptsize aspect term,\scriptsize category,\scriptsize polarity},
                ymin=0,
                ymax=65,
                xmin=-0.5,
                xmax=2.5,
                ymajorgrids=true,
                ylabel={\footnotesize Number of errors},
                xlabel={\footnotesize Sentiment element},
                legend style={at={(1,1)}, anchor=north east, font=\footnotesize},
                ]
                \addplot[black,fill=lightblue,postaction={pattern=north west lines}] coordinates {(0,48) (1,36) (2,9)};
                \addplot[black,fill=lightorange,postaction={pattern=north east lines}] coordinates {(0,39) (1,34) (2,9)};
                \addplot[black,fill=lightgreen,postaction={pattern=grid}] coordinates {(0,55) (1,34) (2,13)};
                \legend{LLaMA 3.1,mT5 w/ CD, mT5 w/o CD}
            \end{axis}
        \end{tikzpicture}
    \end{adjustbox}
    \caption{Number of error types for LLaMA~3.1 and mT5 with and without constrained decoding (CD) on the Spanish target language and the TASD task.}
    \label{fig:error}
\end{figure}
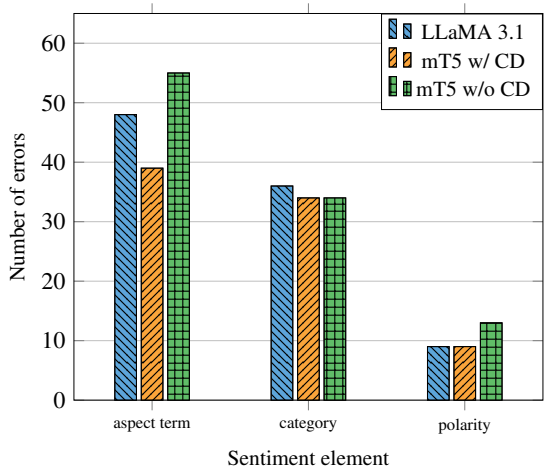

We perform an error analysis to understand the challenges of sentiment prediction better. Specifically, we manually examine the predictions for the first 100 test samples of the Spanish dataset in the TASD task, using the best-performing runs of LLaMA~3.1 and mT5 with and without constrained decoding. Figure~\ref{fig:error} depicts the results of the error analysis.

The primary source of errors lies in aspect term prediction, where the model often misses some aspect terms, generates additional ones, or produces incomplete terms instead of full ones. Without constrained decoding, the mT5 model may also generate text that does not appear in the original review, alter its format, or use the source language instead of the target language (e.g. \textit{\quotes{service}} instead of \textit{\quotes{servicio}}). Constrained decoding addresses these issues effectively, reducing the number of errors.

Further challenges arise from less frequent aspect categories and inconsistent annotations, particularly for categories like \textit{\quotes{restaurant general}} and \textit{\quotes{restaurant miscellaneous}}, which impact performance. Another common error is the confusion between \textit{\quotes{restaurant prices}} and \textit{\quotes{food prices}} categories. Additionally, some categories appear in only one or a few languages, such as \textit{\quotes{food general}}, found exclusively in the Dutch test set, limiting the classifier's ability to learn from other source languages.

Sentiment polarity prediction is generally less challenging than other sentiment elements, with errors primarily occurring in misclassifying the \textit{\quotes{neutral}} polarity.

\section{\texorpdfstring{\uppercase{Conclusion}}{Conclusion}}

This paper addresses three compound cross-lingual ABSA tasks using sequence-to-sequence models and constrained decoding without needing external translation tools. Through extensive experiments in six languages, we emphasize the effectiveness of constrained decoding in enhancing zero-shot cross-lingual ABSA. The proposed approach offers a practical alternative to external translation tools, demonstrating robustness and effectiveness across various language pairs and models, and opens up new possibilities for advanced cross-lingual ABSA. Additionally, we compare our method to modern LLMs, finding that older multilingual models outperform fine-tuned English-centric and closed-source LLMs. We show that fine-tuning multilingual LLMs boosts performance significantly, surpassing smaller models in supported languages.

Future research could explore multi-task learning, where a single model is trained simultaneously on multiple tasks, enabling a unified approach to handling diverse ABSA challenges. Additional experiments could examine various source-target language pair combinations to assess cross-lingual adaptability further. Finally, investigating tasks involving a fourth sentiment element (opinion terms) would be valuable, though current efforts are constrained by the limited availability of annotated data, particularly in languages other than English.

\section*{\uppercase{Acknowledgements}}
This work was created with the partial support of the project R\&D of Technologies for Advanced Digitalization in the Pilsen Metropolitan Area (DigiTech) No. CZ.02.01.01/00/23\_021/0008436 and by the Grant No. SGS-2022-016 Advanced methods of data processing and analysis.
Computational resources were provided by the e-INFRA CZ project (ID:90254), supported by the Ministry of Education, Youth and Sports of the Czech Republic.

\bibliographystyle{apalike}
{\small \bibliography{bibliography}}

\appendix
\section{\texorpdfstring{\uppercase{Constrained Decoding}}{Constrained Decoding}}
\label{appendix:cd}
Algorithm~\ref{algo:constrained} shows the pseudo-code of proposed constrained decoding algorithm.

\begin{algorithm}[ht!]
\caption{Proposed constrained decoding for the TASD task.}
\KwData{Generated sequence, Input sentence tokens, Special token map}
\KwResult{Candidate tokens for the next step}
Get positions of \quotes{[} and \quotes{]} in the generated sequence\;

\If{no \quotes{[} tokens generated}{
    \Return \quotes{[}\;
}

Count \quotes{[} and \quotes{]} tokens and find last \quotes{[}\;

Get last generated token\;

\If{fewer \quotes{]} than \quotes{[} and last generated token is special}{
    \Return \quotes{]}\;
}

\If{last generated token is \quotes{[}}{
    \If{last special token is \quotes{;} or none}{
        \Return \quotes{A}\;
    }
    \If{last special token is \quotes{A}}{
        \Return \quotes{C}\;
    }
    \If{last special token is \quotes{C}}{
        \Return \quotes{P}\;
    }
    \If{last special token is \quotes{P}}{
        \Return \quotes{;}\;
    }
}

\If{last special token is \quotes{;}}{
    \Return \quotes{[}\;
}

Initialize result as an empty list\;

\If{last special token is \quotes{A}}{
    Add input sentence tokens and \quotes{it} to result\;
}

\If{last special token is \quotes{C}}{
    Add category tokens to result\;
}

\If{last special token is \quotes{P}}{
    Add sentiment tokens to result\;
}

\If{last generated token is not \quotes{]}}{
    Add \quotes{]} to result\;
    \If{last special token is \quotes{P}}{
       Add \quotes{$\langle$eos$\rangle$} to result\;
    }
}
\Return result\;
\label{algo:constrained}
\end{algorithm}

\end{document}